\documentclass[conference]{IEEEtran}
\IEEEoverridecommandlockouts
\usepackage{cite}
\usepackage{amsmath,amssymb,amsfonts}
\usepackage{algorithmic}
\usepackage{authblk}
\usepackage{graphicx}
\usepackage{textcomp}
\usepackage{xcolor}
\usepackage{blindtext}
\usepackage[T1]{fontenc}
\usepackage[utf8]{inputenc}
\usepackage{algorithm}
\usepackage{algorithmic}
\usepackage{multirow}
\usepackage{tabularx}
\usepackage{harpoon}
\usepackage{hyperref}
\usepackage{url}
\usepackage{booktabs}
\usepackage{etoolbox}
\usepackage{xkeyval,xcolor}
\usepackage{float}
\usepackage{graphicx}
\usepackage[caption=false]{subfig}

\graphicspath{ {./images/} }
\twocolumn
\title{GPU-Accelerated Mobile Multi-view Style Transfer}
\author{Puneet Kohli}
\author{Saravana Gunaseelan}
\author{Jason Orozco}
\author{Yiwen Hua}
\author{Edward Li}
\author{Nicolas Dahlquist}
\affil{Leia Inc}
\date{\today}

\begin{document}
\maketitle
\section{Abstract}



\textit{An estimated 60\% of smartphones sold in 2018 were equipped with multiple rear cameras, enabling a wide variety of 3D-enabled applications such as 3D Photos. The success of 3D Photo platforms (Facebook 3D Photo, Holopix\texttrademark\ , etc) depend on a steady influx of user generated content. These platforms must provide simple image manipulation tools to facilitate content creation, akin to traditional photo platforms. Artistic neural style transfer, propelled by recent advancements in GPU technology, is one such tool for enhancing traditional photos. However, naively extrapolating single-view neural style transfer to the multi-view scenario produces visually inconsistent results and is prohibitively slow on mobile devices. We present a GPU-accelerated multi-view style transfer pipeline which enforces style consistency between views with on-demand performance on mobile platforms. Our pipeline is modular and creates high quality depth and parallax effects from a stereoscopic image pair.}
\section{Introduction}

Consumers are actively seeking ways to make their traditional photography more immersive and life-like. With the rapidly increasing adoption of multi-camera smartphones, a variety of immersive applications that use depth information are now enabled at mass market scale. One of the most promising applications in this domain is \textit{3D Photos} -- which capture multiple viewpoints of a scene with a parallax effect \cite{kopf2019practical3D}. Coupled with a Lightfield display such as those made by Leia Inc  \cite{fattal2013multi}, 3D Photos are brought to life with added depth \cite{howard1995binocular} and multi-view parallax effects \cite{dogdson2006autostereoscopic}. Platforms such as Facebook 3D Photos and Holopix\texttrademark\ \footnote{An image-sharing social network for multi-view images. https://www.holopix.com} are making it easier for consumers to now capture, edit, and share 3D Photos on a single mobile device. Paramount to the continued success of 3D Photo platforms is the availability of image manipulation and content generation tools that consumers are accustomed to from traditional photo platforms.




To create content for 3D Photo platforms, an image containing information of multiple viewpoints (a \textit{multi-view image}) must be generated. With advances in modern algorithms such as view synthesis \cite{scharstein1996stereo, avidan1997novel, avidan1998novel, martin2008fast, zhou2018mpi} and in-painting \cite{ravi2013image, oh2009hole, telea2004image}, it is now possible to generate multi-view images from stereo pairs even on mobile devices for content creation and manipulation. However, challenges with multi-view content creation still remain. The algorithms involved require large amounts of processing power to perform operations for multiple viewpoints along with higher storage and bandwidth requirements which scales poorly with the number of rendered views. Additionally, mobile platforms have limited computing power to support on-demand (\textasciitilde 0.5 seconds) performance. To enable content creation at scale these challenges must be addressed along with providing users simple tools to generate high quality content.

Artistic style transfers using neural networks, pioneered by Gatys \textit{et. al} \cite{gatys2015neural, gatys2016image}, produce visually pleasing content that has been the topic of research of many recent works \cite{Johnson2016Perceptual, ulyanov2016texture, dumoulin2016learned, sheng2018avatar} and seen widespread success in consumer-facing applications such as Microsoft Pix \cite{hua_msft_pix_2017} and Prisma \cite{prismalabs}. Given the commercial success of neural style transfers as a simple way to make even simple images look artistic, we aim to bring the effect to 3D Photo platforms. On a Lightfield display, seeing the artistic effect come to life due to the parallax and depth effect combined is a truly unique experience. An additional and hidden benefit of applying style transfer to multi-view images is that any artifacts such as edge inconsistency or poor in-painting is generally unnoticable once an artistic style is applied. Though neural style transfer techniques work well on single images, naively applying such techniques on individual views of a multi-view image yields inconsistently styled images which results in 3D fatigue when viewed on a multi-view display \cite{kooi2004visualcomfort}.  

Recent works have aimed to solve the consistency between multiple views for artistic style transfer \cite{chen2018stereoscopic, gong2018neural}. These approaches only consider two views and are not directly extendable to multiple views. They are also optimized for specific styles and need to be re-trained for new styles. Video style transfer \cite{ruder2016artistic, Chen_2017_ICCV, huang2017real} is another similar line of work where consistent results need to be produced in the temporal domain. These techniques are currently not computationally efficient enough for running in real-time on mobile devices. In our case, we aim to solve a generalized scenario where multiple views have an artistic style applied, generated from input stereo image and disparity pairs. Our method is agnostic to both the number of views being generated and the style transfer model being used.
 \begin{figure}[H]
     \begin{tabular}{c@{\hskip 0.04in}c@{\hskip 0.04in}c@{\hskip 0.04in}c@{\hskip 0.04in}c}
     \includegraphics[width = 0.63in]{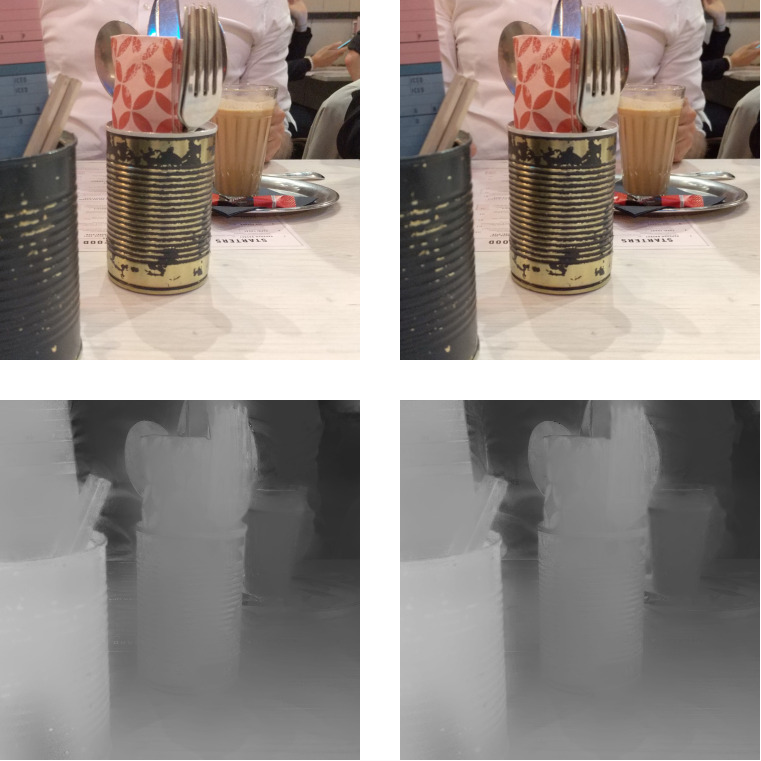} &
    \includegraphics[width = 0.63in]{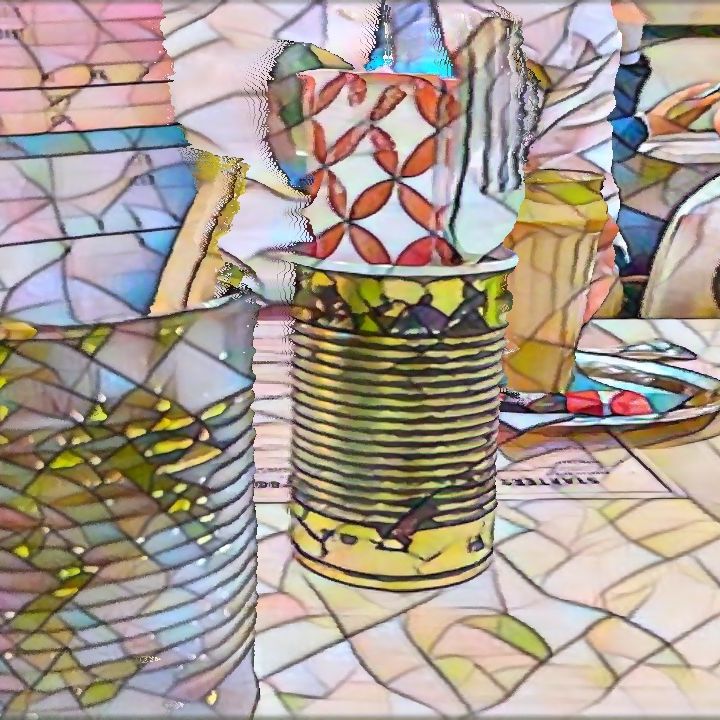} &
    \includegraphics[width = 0.63in]{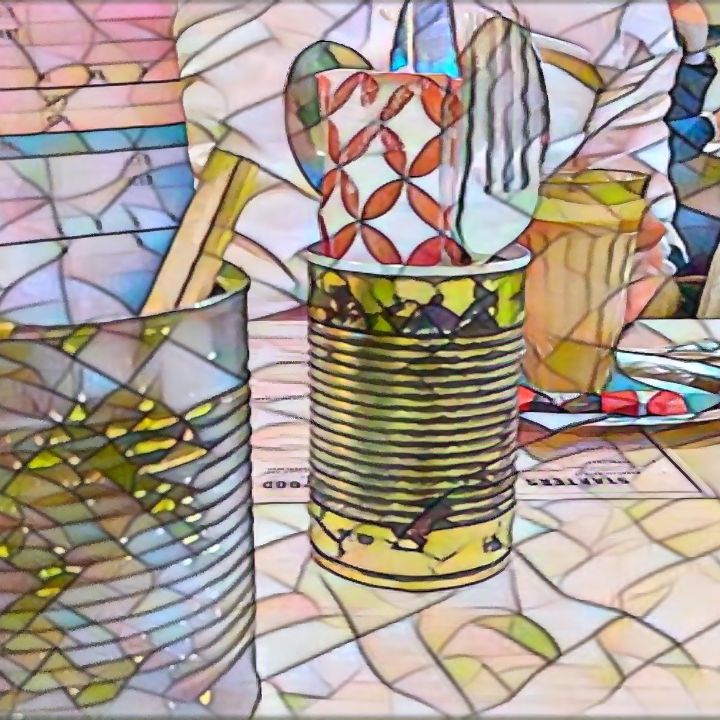} &
    \includegraphics[width = 0.63in]{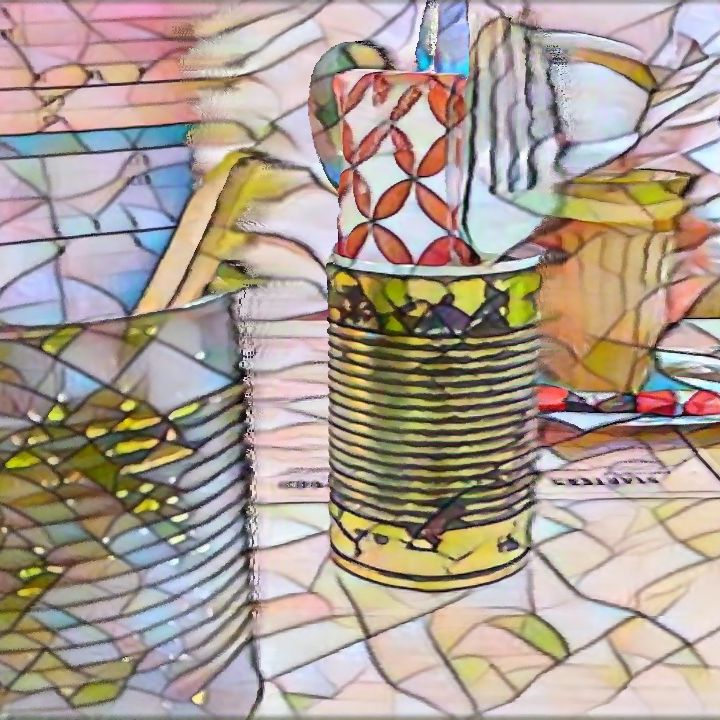} &
    \includegraphics[width = 0.63in]{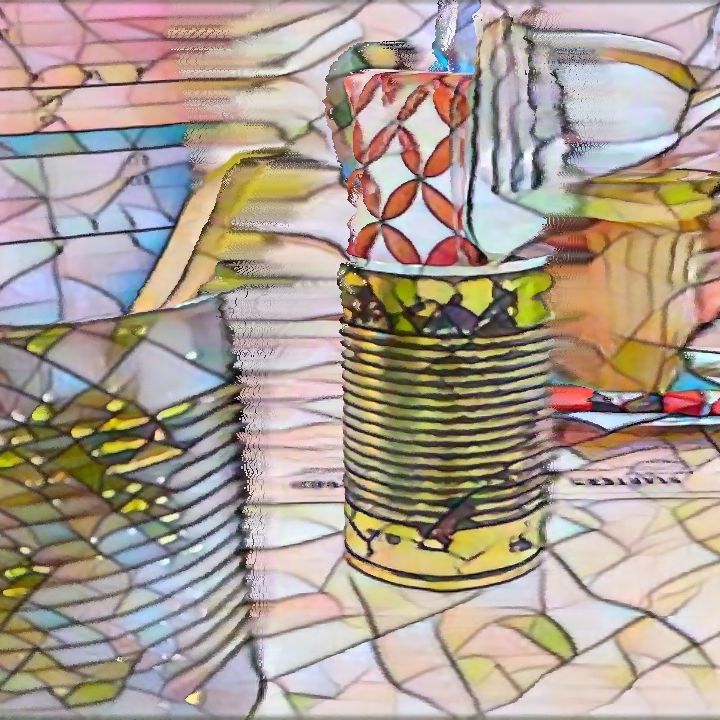}
    \end{tabular}
\caption{An example result from running our style transfer pipeline to generate four views using a stereo image and disparity pair.}
    \label{fig:style_fig}
\end{figure}
To the best of our knowledge, there are no techniques that maintain multi-view style consistency while also addressing the performance concerns on mobile devices. In this work, we propose an end-to-end pipeline for generating stylized multi-view imagery given stereo image and disparity pairs that can run on-demand even on mobile computing platforms. We address the issues of both multi-view style consistency as well as performance.  An example result from our pipeline is shown in Figure \ref{fig:style_fig}.

The proposed method is highly configurable and can support different algorithms for each of the steps involved. This includes style transfer, novel view synthesis \cite{scharstein1996stereo, avidan1997novel, avidan1998novel, martin2008fast, zhou2018mpi}, and in-painting \cite{ravi2013image, oh2009hole, telea2004image}. In a highly modular fashion, each component of the pipeline is independent of each other and can be replaced by an equivalent algorithm. This facilitates fine-tuning the proposed pipeline for different design constraints such as optimizing for power or quality.  

%

In summary, the pipeline proposed in this work has the following main contributions: 

\begin{itemize}
\item Multi-view Consistent neural style transfer and parallax effects 
\item GPU-Accelerated on-demand performance (\textasciitilde0.5s) on mobile platforms. 
\item Modular components for individual algorithms.
\end{itemize}

\section{Methodology} \label{methodology}
\begin{figure}
    \centering
    \includegraphics[width=0.5\textheight,height=0.5\textheight,keepaspectratio]{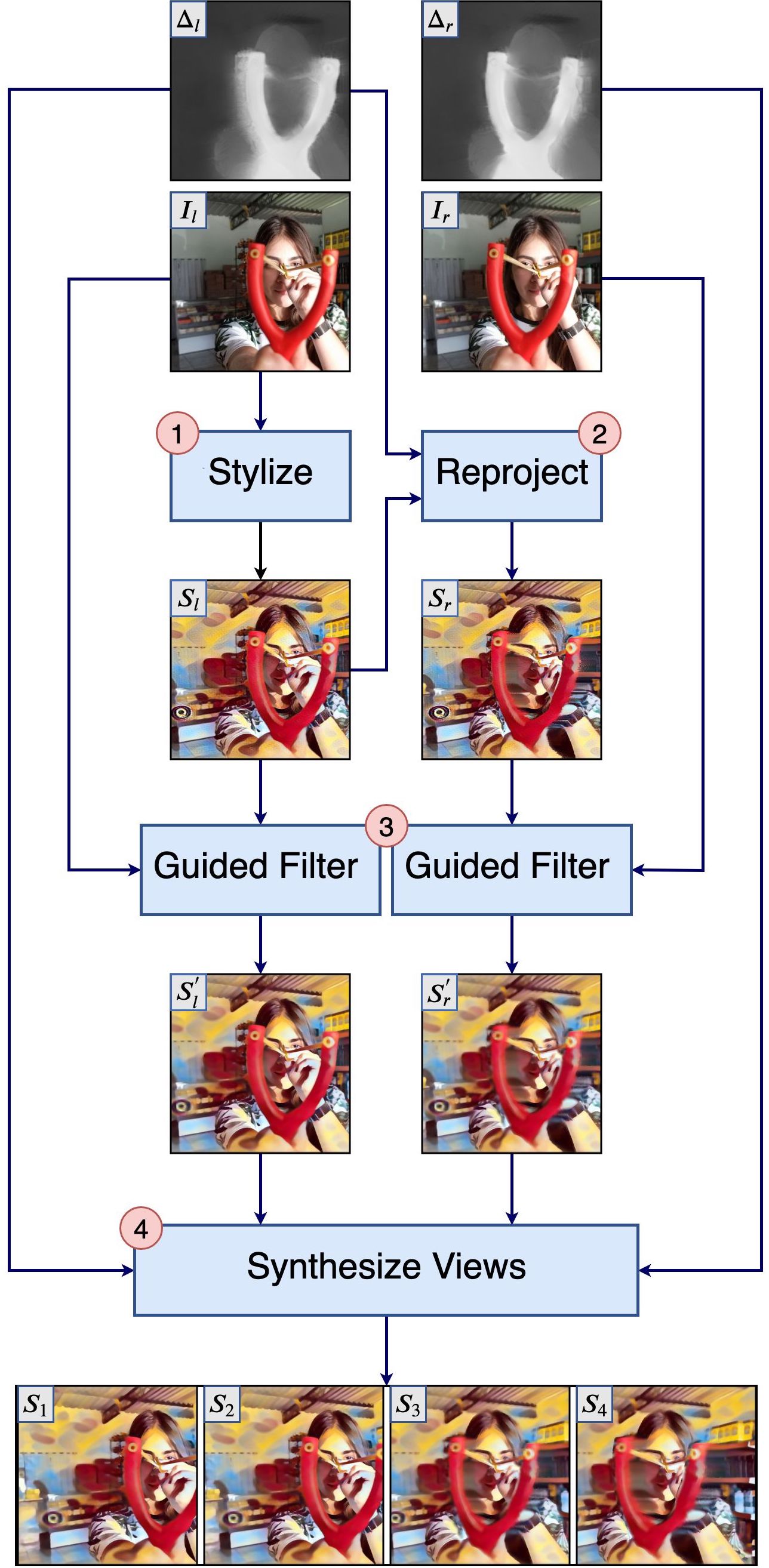}
    \caption{Algorithm flow of our multi-view style transfer pipeline. Red numbers correspond to steps in Algorithm \ref{alg:alg1}.}
    \label{fig:alg_flow}
\end{figure}

The proposed method for multi-view consistent stylizing combines existing monoscopic style transfer techniques with view synthesis in a highly modular fashion, while promoting 3D scene integrity.  An overview is given in Algorithm \ref{alg:alg1} and Figure \ref{fig:alg_flow} showing the four steps to stylize the left view, re-project to the right viewpoint, apply a guided filter to the left and right stylized views, then synthesize any number of stylized novel views. This section explores how each module contributes to create immersive stylized 3D content.

\begin{algorithm}[H]
\caption{Multi-view consistent style transfer. Given stereo input views $I_l,I_r$, disparity maps $\Delta_l,\Delta_r$, and style guide $G_s$, render stylized views $S_{1,2,...,n}$ at desired output viewpoints $x_1,x_2,...,x_n$.}
\label{alg:alg1}
\begin{algorithmic}[1]
\STATE Infer stylized left view $S_l$ from $I_l$ and $G_s$ using style transfer module.
\STATE Re-project $S_l$ to $S_r$ at the same viewpoint as $I_r$ using view synthesis module.
\STATE Apply guided filter module to $S_l,S_r$ with guides $I_l,I_r$ to produce filtered stylized views $S_l',S_r'$.
\FOR {$i$ in \{1,2,...,n\}} 
    \STATE Re-project $S_l'$ and $S_r'$ to $x_i$ using view synthesis module and blend into $S_i$.
\ENDFOR
\end{algorithmic}
\end{algorithm}

\subsection{Input}
We assume that left and right views $I_l,I_r$ are given with corresponding disparity maps $\Delta_l,\Delta_r$.  In applications where disparity maps are not given, a variety of well-known estimation methods can be used \cite{hoff1989surfaces, wanner2013variational, mayer2016large, luo2016efficient}. In our case, we use a neural network trained on stereo images from Holopix\texttrademark\ prior to our experiments.

\subsection{Style Transfer}

The proposed method stylizes the left view $S_l$ using existing style transfer methods \cite{zhang2017multistyle, Johnson2016Perceptual}. We use a network based on  \cite{Johnson2016Perceptual} along with Instance Normalization \cite{ulyanov2016instance} in our experiments.

We achieve three main benefits by applying the style transfer network to stylize only one of the input stereo views. The first is to achieve style consistency between multiple views, which is the primary motivation for our proposed method. The challenge is that  style transfer networks are sensitive to small changes in the input. When stylizing each view individually, the parallax effect between neighboring views is enough to cause visual differences such as stylized features entirely appearing and disappearing from one view to the next (see Figure \ref{fig:detailed_comparison}). Such inconsistencies cause 3D fatigue when viewed on a multi-view display.

Second, running a style transfer network incurs a significant performance cost even with competitive technology, so stylizing once and synthesizing many is much faster than stylizing every output view individually.

The third benefit is compatibility with any style transfer network. By not demanding modifications and retraining specifically for multi-view rendering, any network that takes a single image and returns a stylized version can be readily substituted in our pipeline, provided the quality and performance are satisfactory for single images.

\subsection{One In, One Out View Synthesis}
\label{oneinoneout}
Next, the stylized left view $S_l$ is re-projected to $S_r$ at the same viewpoint as the input right view $I_r$ using a view synthesis module. View synthesis is an active area of research with many existing solutions \cite{scharstein1996stereo, avidan1997novel, avidan1998novel, martin2008fast}. We use an in-house algorithm for performing view synthesis both on the CPU and GPU. Our algorithm incorporates forward warping with a depth test and a simple in-painting technique that samples nearby regions to fill deoccluded regions.

The effect of re-projecting the stylized left view to the right viewpoint is that the style features are transported precisely to their corresponding positions in the right view. This is equivalent to having the style features present in the 3D scene at the time the stereo photo is taken.

\subsection{Guided Filter}

\begin{figure}
    \centering
    \includegraphics[width=0.35\textheight,height=0.35\textheight,keepaspectratio]{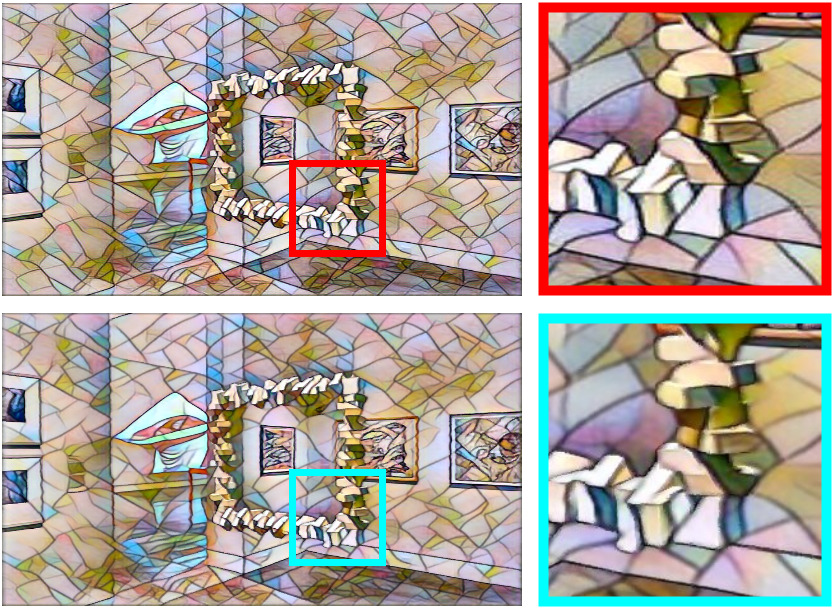}
    \caption{Comparison of stylized right image before (top row) and after (bottom row) applying guided filter. The highlighted regions show that applying a guided filter emphasizes the edges in the original 3D scene over the edges introduced by style transfer.}
    \label{fig:guided_style}
\end{figure}

The stylized views are refined by the edge-aware guided filtering methodology to produce filtered stylized left and right views $S_l',S_r'$ with the input views $I_l,I_r$ serving as their respective guides. We use the CPU-based guided filter \cite{He:2013:GIF:2498746.2498975} in all our experiments.

When viewing images on a multi-view display, the quality of edges plays an essential role in 3D perception, however, the style transfer process tends to degrade the edges of objects in the 3D scene. By applying a guided filter to the stylized views using their corresponding un-stylized views as guides, the edges of original 3D objects are reinforced while reducing the stylization edges, resulting in a more immersive 3D experience. See Figure \ref{fig:guided_style} for comparison.

\subsection{Two In, Many Out View Synthesis}

The final step is to synthesize the desired output views $S_{1,2,...,n}$ from the filtered stylized left and right views $S_l',S_r'$ and given disparity maps $\Delta_l, \Delta_r$.  This is done by repeated application of the same \emph{one in, one out view synthesis} module described in Section \ref{oneinoneout} to re-project both $S_l'$ and $S_r'$ to every output viewpoint. Each output view $S_i$ is the result of re-projecting both $S_l'$ and $S_r'$ to the desired viewpoint of $S_i$, say $x$, and blending based on proximity of $x$ to the left and right viewpoints $l$ and $r$.

\section{Experiments and Results}

\subsection{Experimentation}

We tested four approaches for generating multi-view style transferred images and compared both qualitative and quantitative results. We ran all experiments on a RED Hydrogen One \cite{redhydrogen} mobile phone that has four-view autostereoscopic display. The device runs on the Qualcomm® Snapdragon 835 Mobile Platform coupled with the Qualcomm® Adreno™ 540 GPU. 
\\\\
\textbf{Baseline Approach.} The first approach naively applies neural style transfer to each of the synthesized views individually. This fails to produce style consistent views due to the unstable nature of neural style transfer models \cite{chen2018stereoscopic} and also does not produce on-demand results. A supplementary observation shows that style transfer is considerably slower than view synthesis. The overall performance scales poorly with the number of views.
\\\\
\textbf{Approach 2.} Neural style-transfer is first applied to each of the stereoscopic input views and then novel view synthesis is performed using the stylized pair and the original disparity maps as inputs. Although this runs significantly faster than the \textit{baseline} method, the rendered views produce undesirable ghosting artifacts and an overall inconsistent styling between the stereoscopic pair leading to 3D fatigue. 
\\\\
\textbf{Approach 3.} Here the style inconsistency problem left unsolved from the previous methods is tackled. Neural style is applied only to the input left image to create the stylized left image. Novel views are synthesized only from this stylized left image. View synthesis is simultaneously performed using both the original naturalistic left and right images. This naturalistic multi-view image is used as a guide for a guided filter pass on the stylized multi-view image. The resulting multi-view image is sharpened to reduce blurring artifacts. This method produces consistently styled views with relatively sharp edges. 
\\
However, the drawback of this approach is that there is a limited depth effect due to using only the left image for the styled novel view synthesis. Additionally, as the edges do not line up perfectly in the guided filtering step, ghosting artifacts are produced around the edges in the output views.
\\\\
\textbf{Selected Approach.} Our novel approach outlined in Section \ref{methodology} builds upon ideas from all of the previous approaches and succeeds in producing multi-view consistent stylized images with on-demand performance when GPU-accelerated on mobile devices. 

\subsection{Quantitative Evaluation} 

Our evaluation metric for comparing the various approaches is the time taken to produce a multi-view image with neural style transfer applied from an input stereoscopic image pair and their associated disparity maps. We run each of the compared methods on the \textit{MPI Sintel} stereo test dataset\cite{butler2012sintel} at a 1024x436 resolution.
\\


\begin{table}[ht]
\renewcommand{\arraystretch}{1.3}
\caption{Mean runtime comparison for rendering a stylized multi-view image with 4, 8, and 16 views from the MPI Sintel stereo dataset.}
\centering
\begin{tabular}[t]{lccc}
\toprule
\multirow{2}{*}{Method} & 
\multicolumn{3}{c}{Time taken (ms)} \\
& 4 Views & 8 Views & 16 Views \\
\midrule
Baseline$_{CPU}$ & 8352 & 16682 & 33358 \\
Baseline$_{GPU}$ & 1405 & 2832 & 5768 \\
Approach$_{2}$ & 843 & 849 & 858 \\
Approach$_{3}$ & 746 & 995 & 1213 \\
Ours$_{CPU}$ & 2311 & 2394 & 2576 \\
\textbf{Ours$_{GPU}$} & \textbf{561} & \textbf{567} & \textbf{576} \\
\bottomrule
\label{tab:results_rh1}
\end{tabular}
\end{table}

Table \ref{tab:results_rh1} shows the run time comparison of our various methods on the \textit{MPI Sintel} stereo dataset. The baseline method is run both in an end-to-end CPU and a GPU-accelerated environment to highlight the performance advantages of using the GPU. As can be seen, our novel pipeline is the fastest method and scales linearly with the number of views generated. The style transfer model used in our experiments is from \cite{Johnson2016Perceptual}.
\\

Additionally, we deployed images with our stylization technique to the Holopix\texttrademark\ platform and compared overall engagement. We observed that images posted to Holopix\texttrademark\ with style transfer applied perform over 300\% better than the platform average. In a control post when the same image was posted twice in succession, once with style transfer applied, and once without, the version of the image with style transfer performed over 20\% better.

\subsection{Qualitative Evaluation}
\begin{figure}
    \centering
    \includegraphics[width=0.35\textheight,height=0.35\textheight,keepaspectratio]{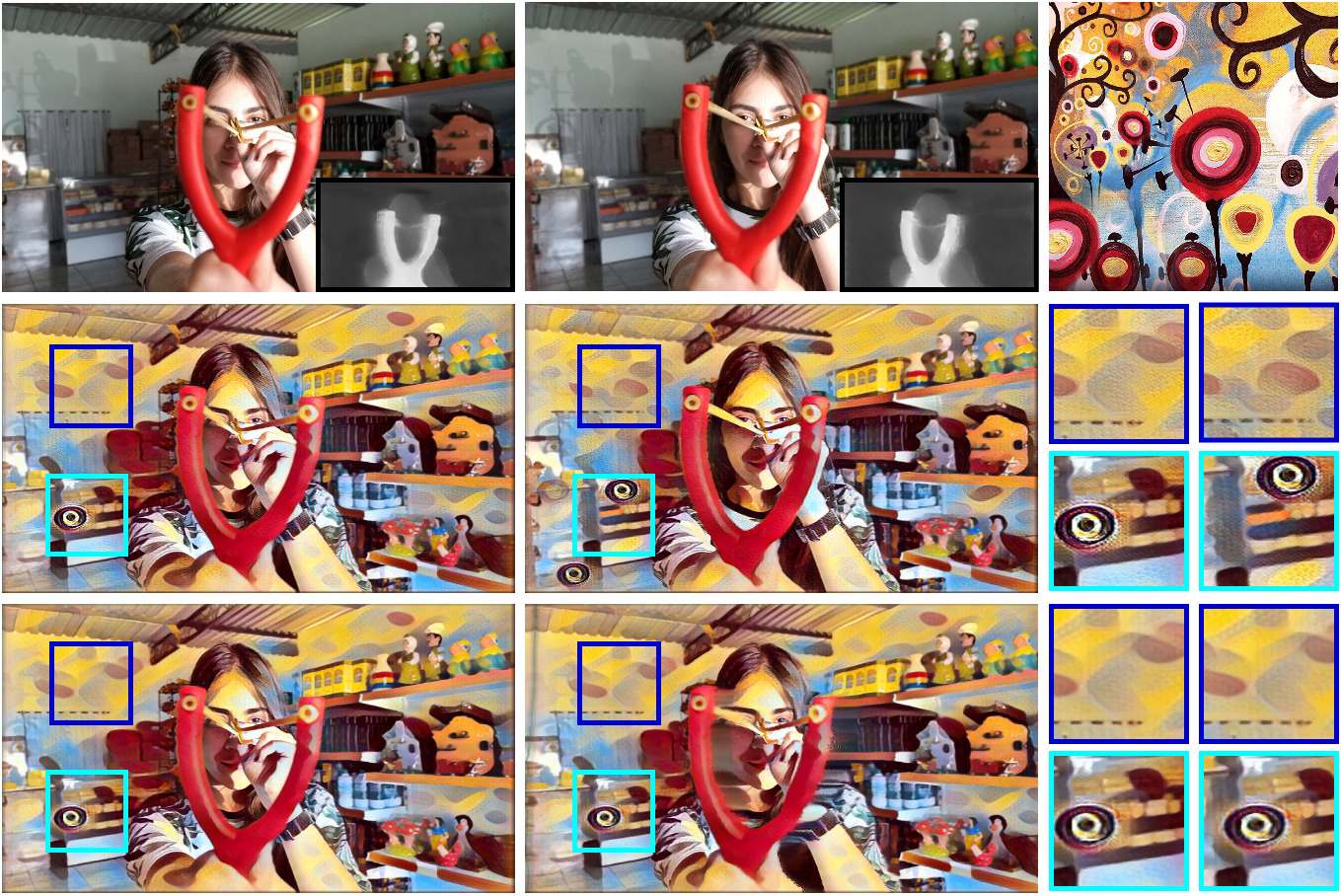}
    \caption{Top row: original left image and disparity (left), original right image and disparity (center), style guide (right). Comparison of stylized left and right image using baseline approach (middle) and our approach (bottom). Baseline approach produces inconsistent styling between left and right image leading to 3D fatigue and our approach eliminates such inconsistency as seen in the highlighted region.
}
    \label{fig:detailed_comparison}
\end{figure}

Figure \ref{fig:detailed_comparison} shows visual results from the baseline approach vs our final approach for a stereoscopic image pair taken from a stereo camera on the RED Hydrogen One \cite{redhydrogen} device. When viewing the results on a multi-view display, the baseline methods makes it difficult for the eyes to focus on due to inconsistencies causing 3D fatigue. In comparison, our approach has consistent views making the experience more comfortable to view on a multi-view display. 

We show additional results in Figure \ref{fig:more_results} where we generate various images from stereo pairs and a different style. It is interesting to note that between the first view of each result and the last view, objects have consistent stylization despite having different positions, rotations, or occlusions.

            

\section{Conclusion}

 In  this  work, we  propose  an  end-to-end  pipeline  for  generating  stylized multi-view imagery given stereo image and disparity pairs. Our experiments show that the pipeline can achieve on-demand (\textasciitilde 0.5 seconds) results on mobile GPUs, tested up to sixteen views. The results are style consistent between each of the views and produce a highly immersive parallax effect when viewed on a lightfield display. Finally, the pipeline is modular and can support any style transfer, view synthesis, or in-painting algorithm. To the best of our knowledge, this is the first published work for producing style consistent multi-view imagery from a stereoscopic image and disparity pair with on-demand performance on mobile GPUs. In a future work, we will run our pipeline on the Jetson TX2 GPU and the Tegra X1 GPU on an NVIDIA SHIELD and we expect to see similar or better quantitative results.
 \newcommand{\subfigimg}[3][,]{%
  \setbox1=\hbox{\includegraphics[#1]{#3}}
  \leavevmode\rlap{\usebox1}
  \setbox2=\hbox{\includegraphics[width=0.15in]{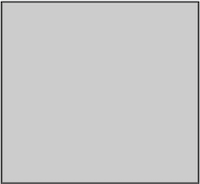}}
  \rlap{\hspace*{-3.5pt}\raisebox{\dimexpr\ht1-10pt}{\usebox2}}
  \rlap{\hspace*{-1pt}\raisebox{\dimexpr\ht1-8pt}{\color{black}{#2}}}
  \phantom{\usebox1}
}

\newcommand{\subfigimgs}[3][,]{%
  \setbox1=\hbox{\includegraphics[#1]{#3}}
  \leavevmode\rlap{\usebox1}
  \setbox2=\hbox{\includegraphics[width=0.15in]{images/more_results/box.png}}
  \rlap{\hspace*{-3.5pt}\raisebox{\dimexpr\ht1-10pt}{\usebox2}}
  \rlap{\hspace*{-3.5pt}\raisebox{\dimexpr\ht1-8pt}{\color{black}{#2}}}
  \phantom{\usebox1}
}

\makeatletter
\setlength{\@fptop}{0pt}
\makeatother
\begin{figure}

    \begin{tabular}{c@{\hskip 0.06in}c@{\hskip 0.06in}c@{\hskip 0.06in}c@{\hskip 0.06in}c@{\hskip 0.06in}c}
    \centering
    \subfloat{\includegraphics[width = 0.75in, height = 0.345in]{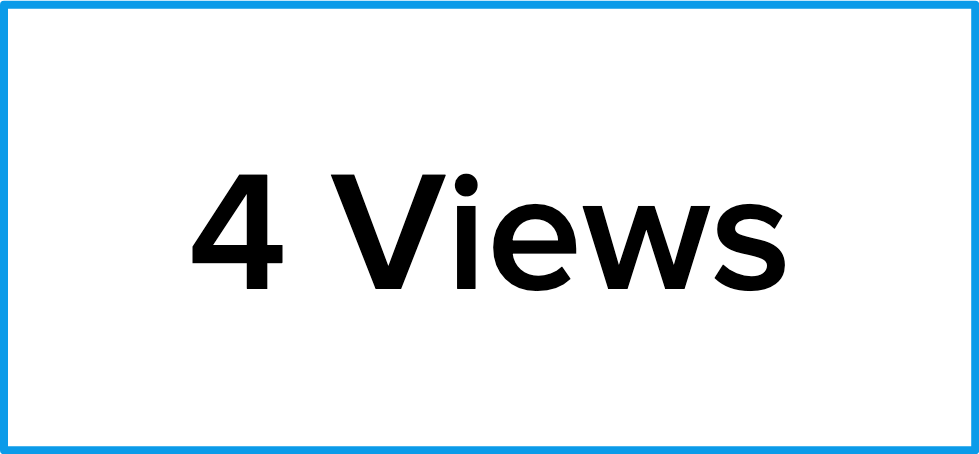}}
    \subfloat{\includegraphics[width = 0.345in]{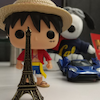}}&
    \subfloat{\includegraphics[width = 0.345in]{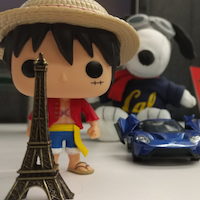}} &
    \subfloat{\includegraphics[width = 0.345in]{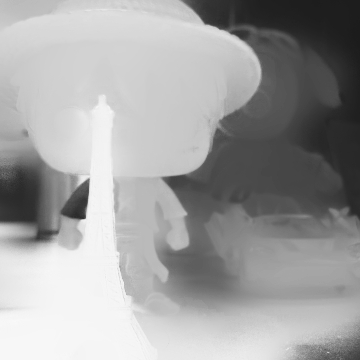}} &
    \subfloat{\includegraphics[width = 0.345in]{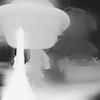}} &
    \subfloat{\includegraphics[width = 0.345in, height = 0.345in]{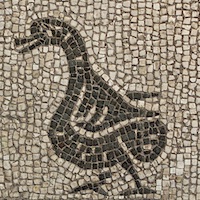}}
    \end{tabular}
    \begin{tabular}{c@{\hskip 0.01in}c@{\hskip 0.01in}c@{\hskip 0.01in}c}
    \subfigimg[width = 0.75in]{1}{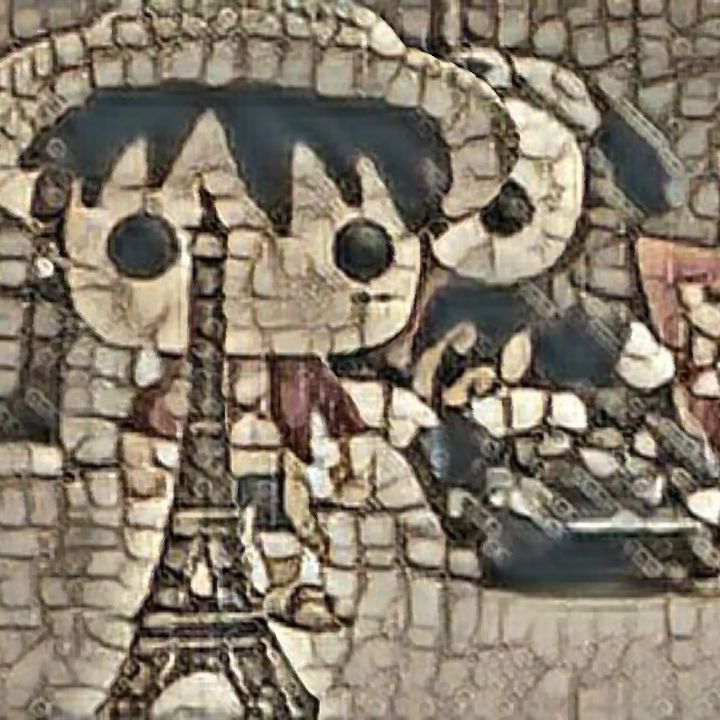} &
    \subfigimg[width = 0.75in]{2}{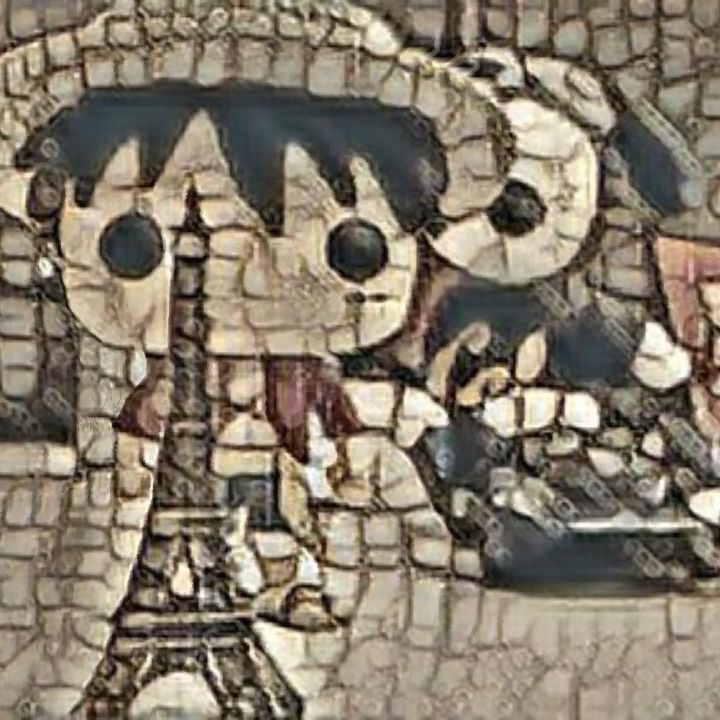} &
    \subfigimg[width=0.75in]{3}{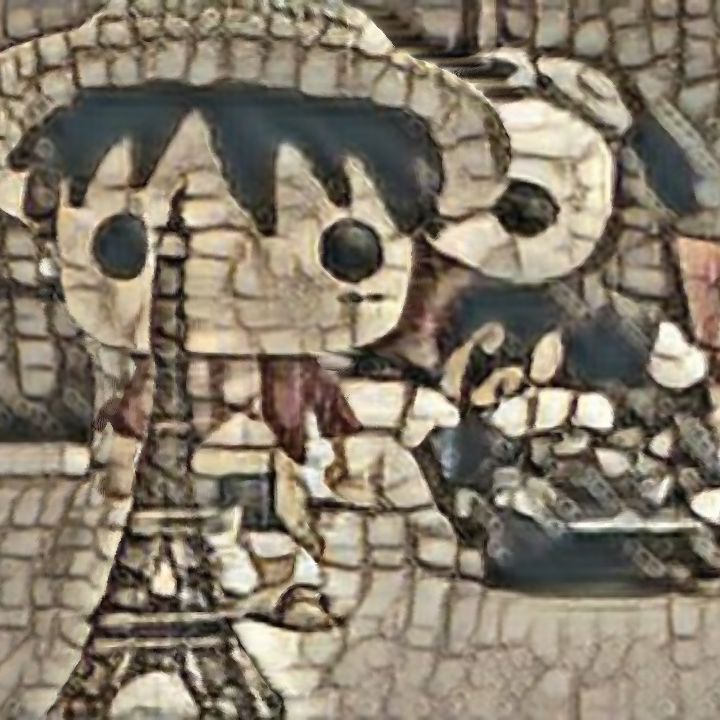} &
    \subfigimg[width = 0.75in]{4}{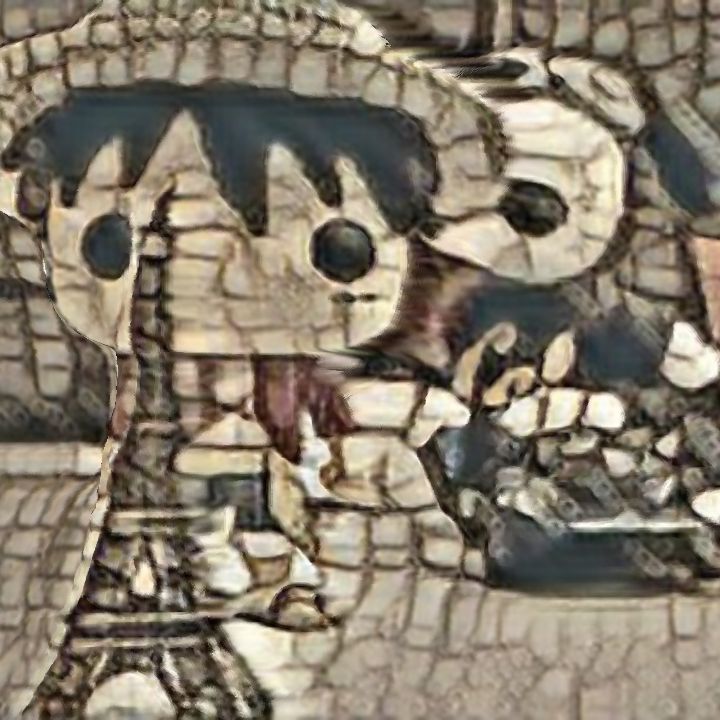}
    \end{tabular}
    \newline
    \vspace*{0.15in}
    \newline
    \begin{tabular}{c@{\hskip 0.06in}c@{\hskip 0.06in}c@{\hskip 0.06in}c@{\hskip 0.06in}c@{\hskip 0.06in}c}
    \subfloat{\includegraphics[width = 0.75in, height = 0.345in]{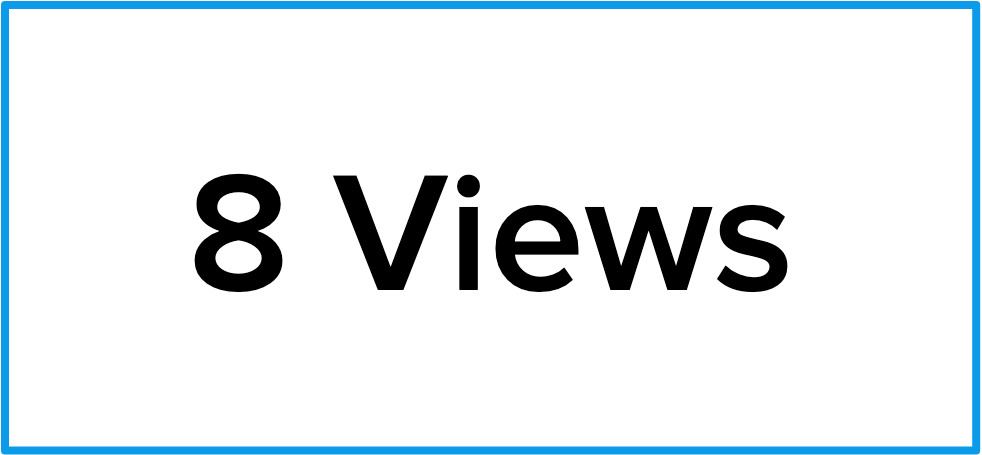}} &
    \subfloat{\includegraphics[width = 0.345in]{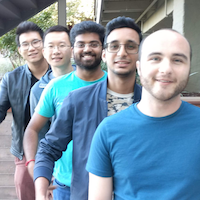}} &
    \subfloat{\includegraphics[width = 0.345in]{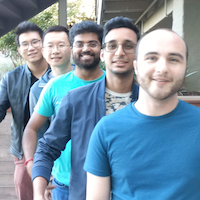}} &
    \subfloat{\includegraphics[width = 0.345in]{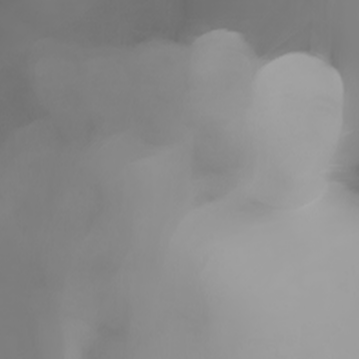}} &
    \subfloat{\includegraphics[width = 0.345in]{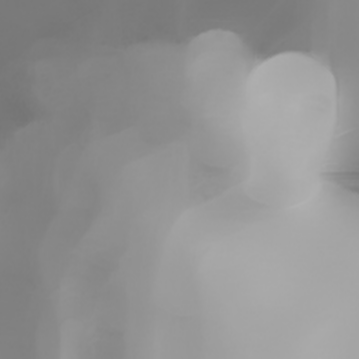}} &
    \subfloat{\includegraphics[width = 0.345in, height = 0.345in]{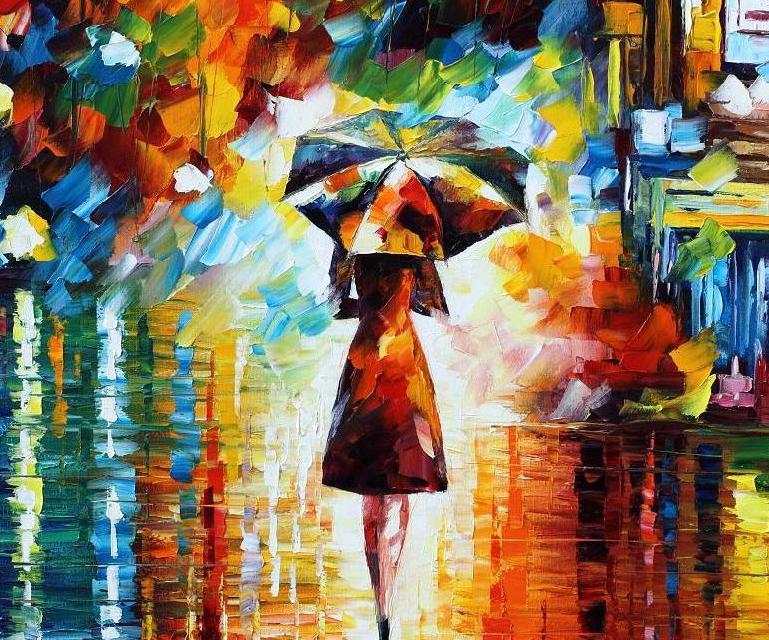}}
    \end{tabular}
    \begin{tabular}{c@{\hskip 0.01in}c@{\hskip 0.01in}c@{\hskip 0.01in}c}
    \subfigimg[width = 0.75in]{1}{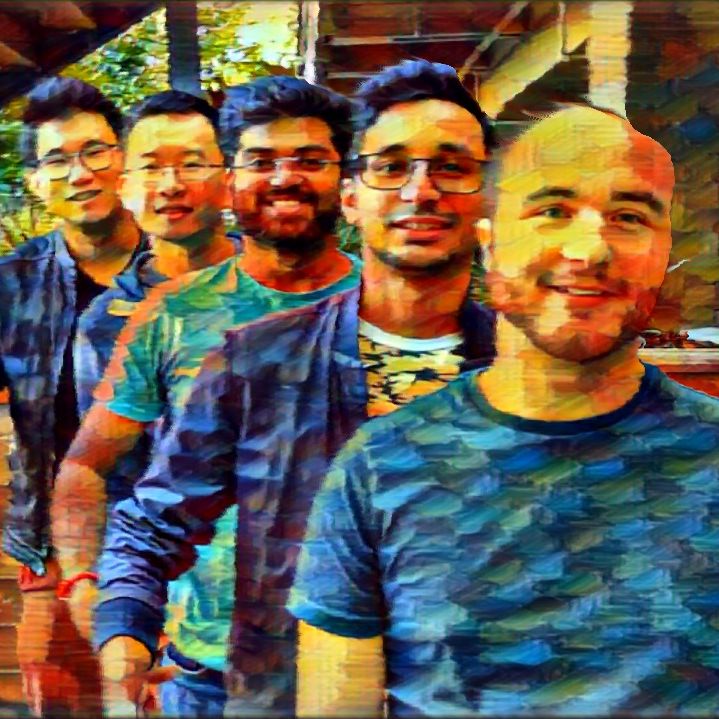} &
    \subfigimg[width = 0.75in]{2}{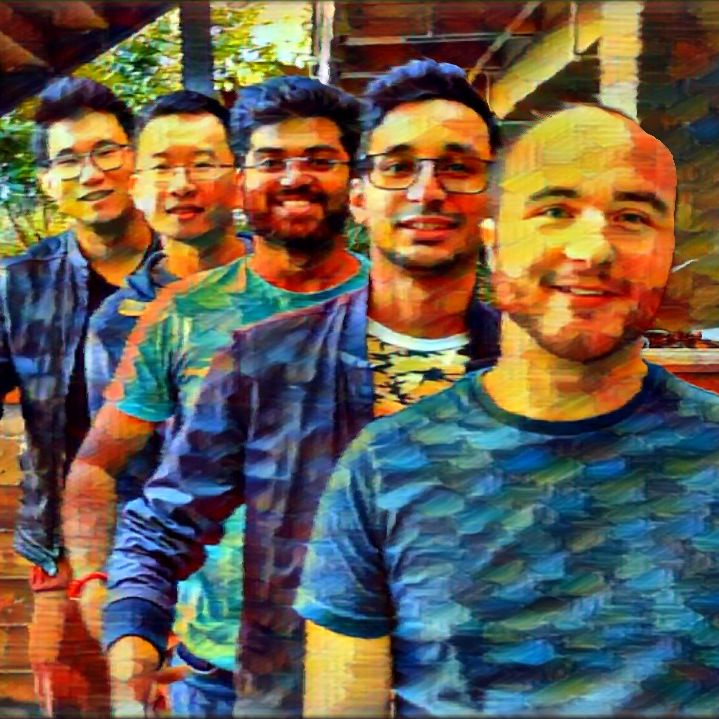} &
    \subfigimg[width=0.75in]{3}{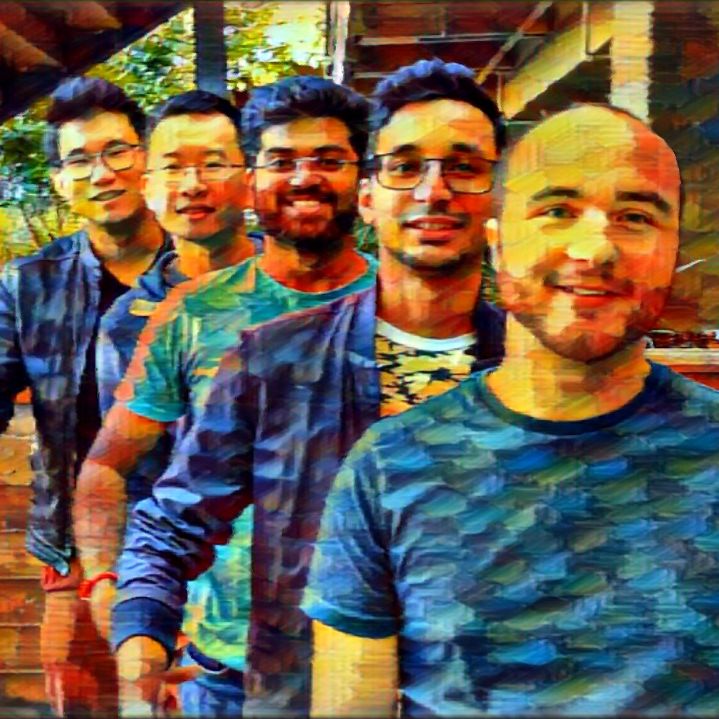} &
    \subfigimg[width = 0.75in]{4}{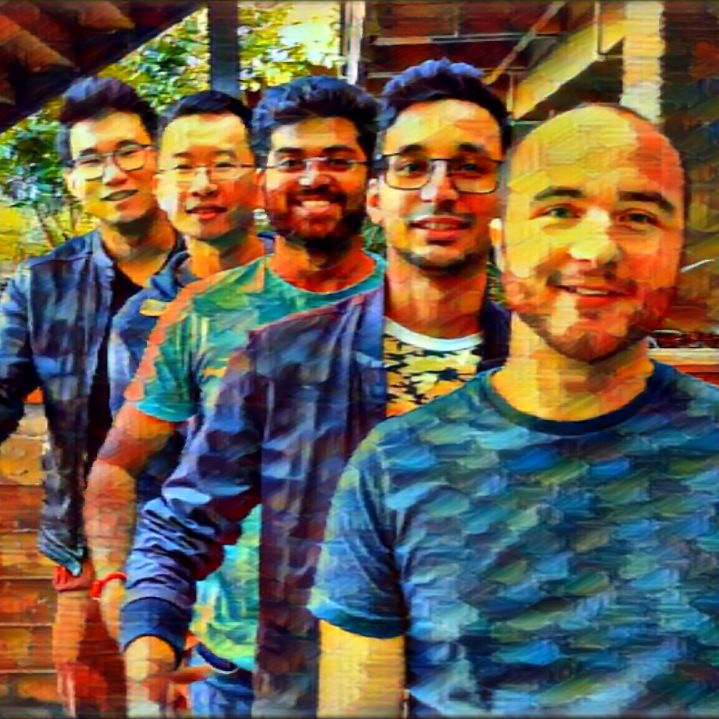}\\
    \subfigimg[width = 0.75in]{5}{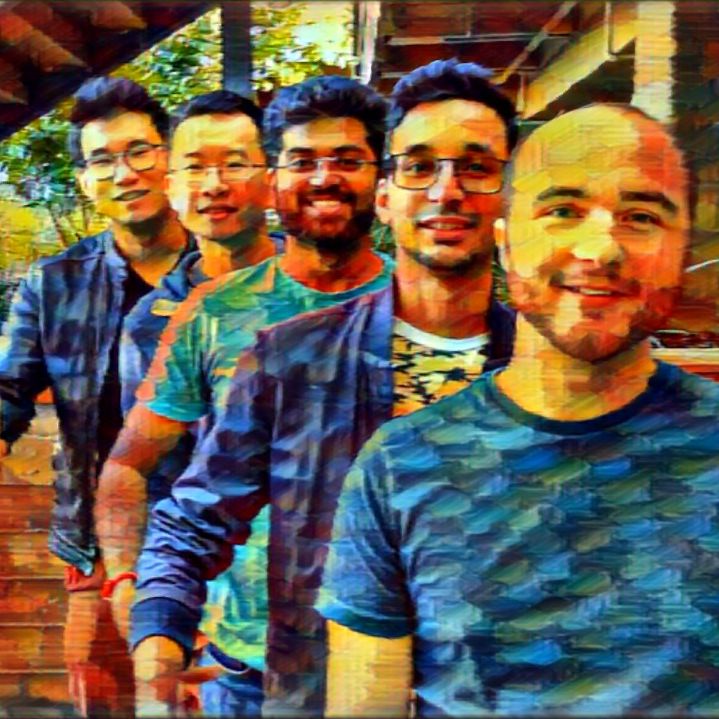} &
    \subfigimg[width = 0.75in]{6}{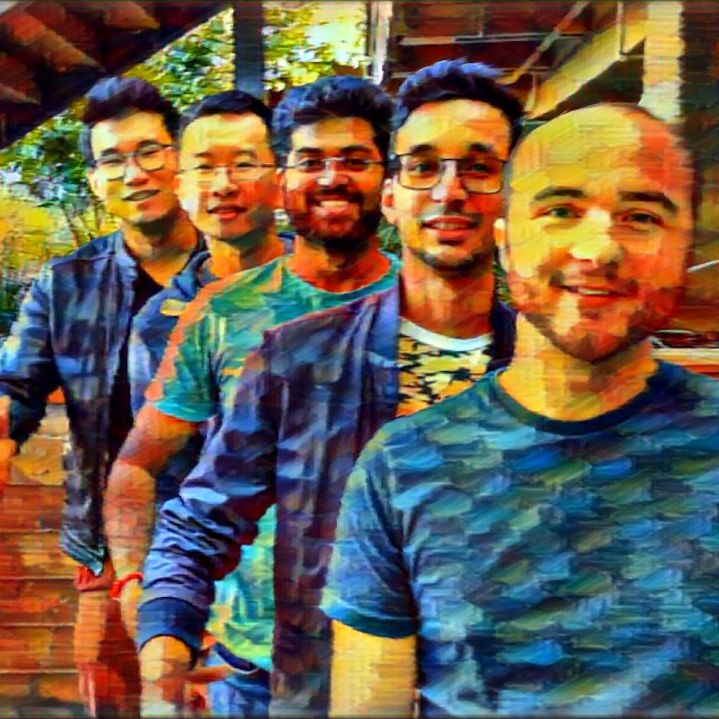} &
    \subfigimg[width = 0.75in]{7}{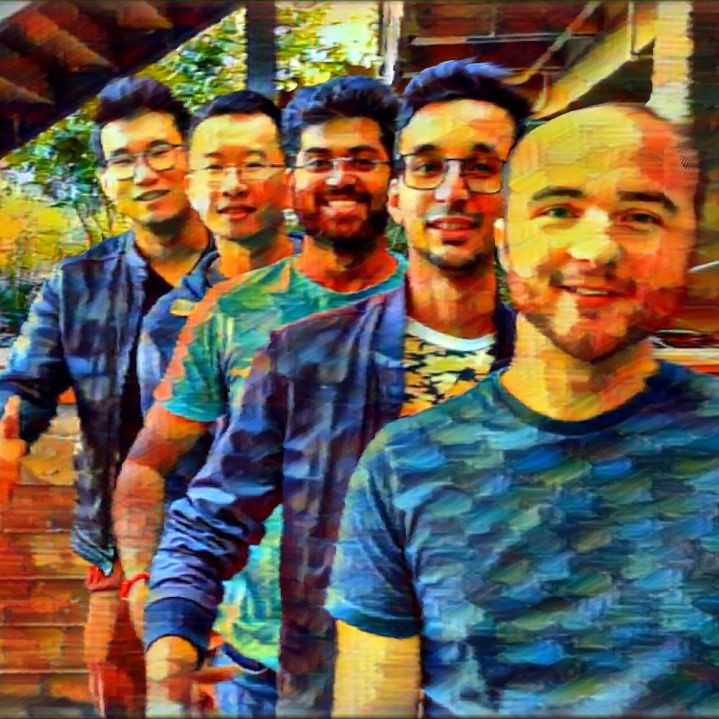} &
    \subfigimg[width = 0.75in]{8}{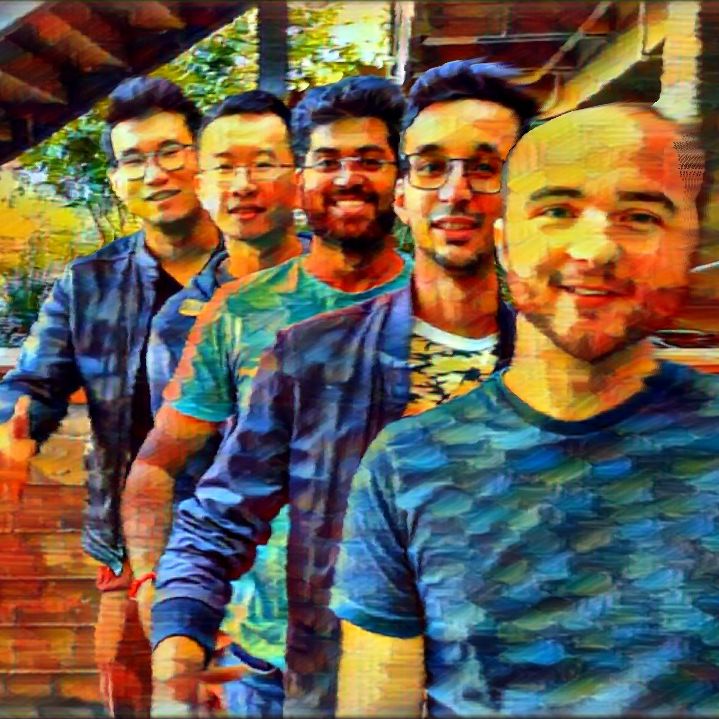}
    \end{tabular}
    \newline
    \vspace*{0.15in}
    \newline
    \begin{tabular}{c@{\hskip 0.06in}c@{\hskip 0.06in}c@{\hskip 0.06in}c@{\hskip 0.06in}c@{\hskip 0.06in}c}
    \subfloat{\includegraphics[width = 0.75in, height = 0.345in]{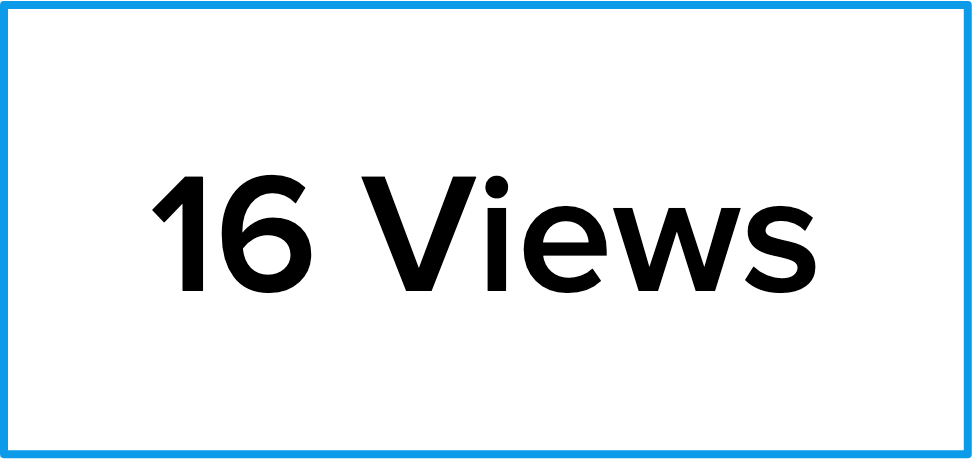}} &
    \subfloat{\includegraphics[width = 0.345in]{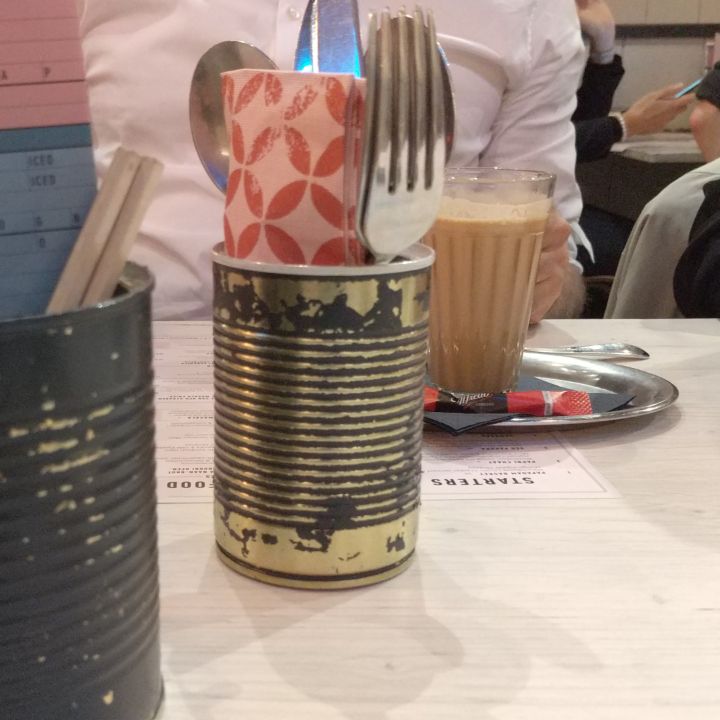}} &
    \subfloat{\includegraphics[width = 0.345in]{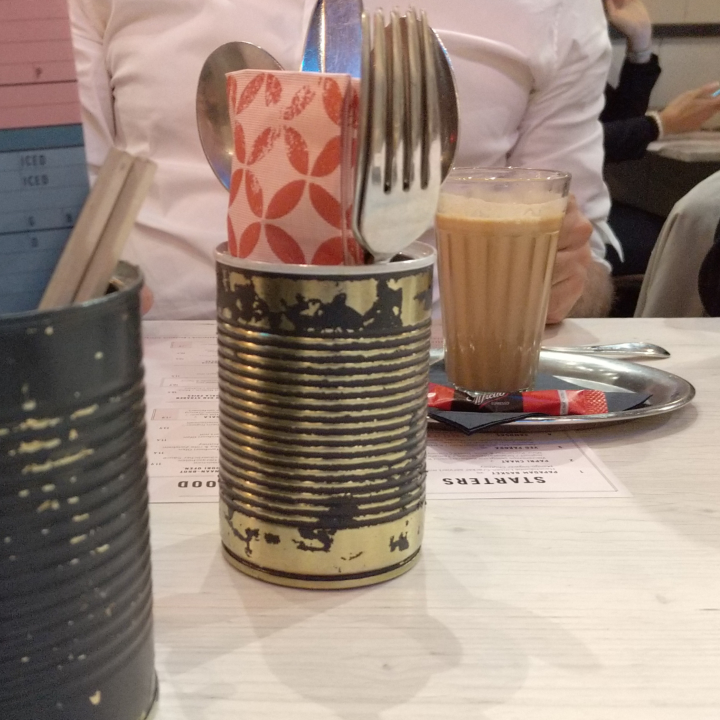}} &
    \subfloat{\includegraphics[width = 0.345in]{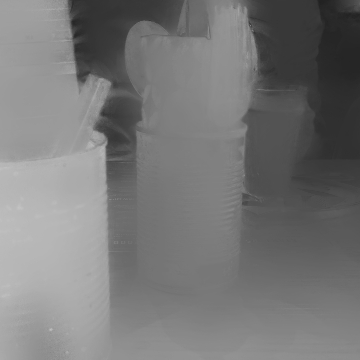}} &
    \subfloat{\includegraphics[width = 0.345in]{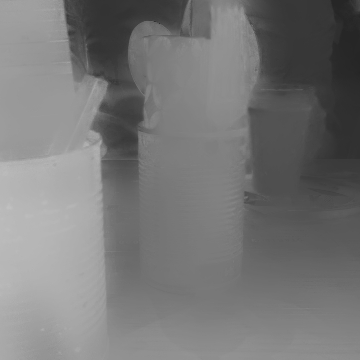}} &
    \subfloat{\includegraphics[width = 0.345in, height = 0.345in]{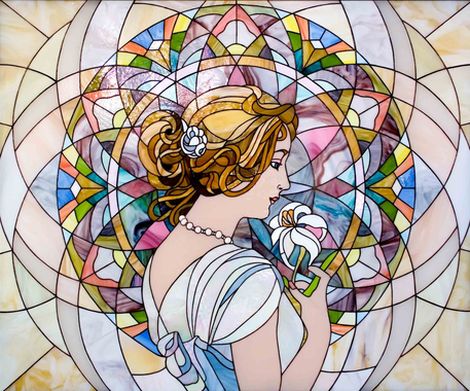}}
    \end{tabular}
    \begin{tabular}{c@{\hskip 0.01in}c@{\hskip 0.01in}c@{\hskip 0.01in}c}
    \subfigimg[width = 0.75in]{1}{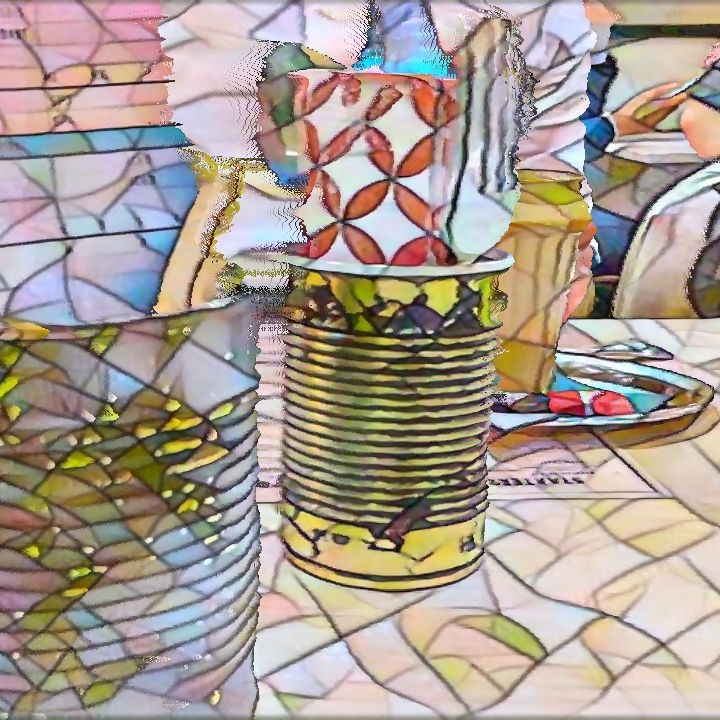}&
    \subfigimg[width = 0.75in]{2}{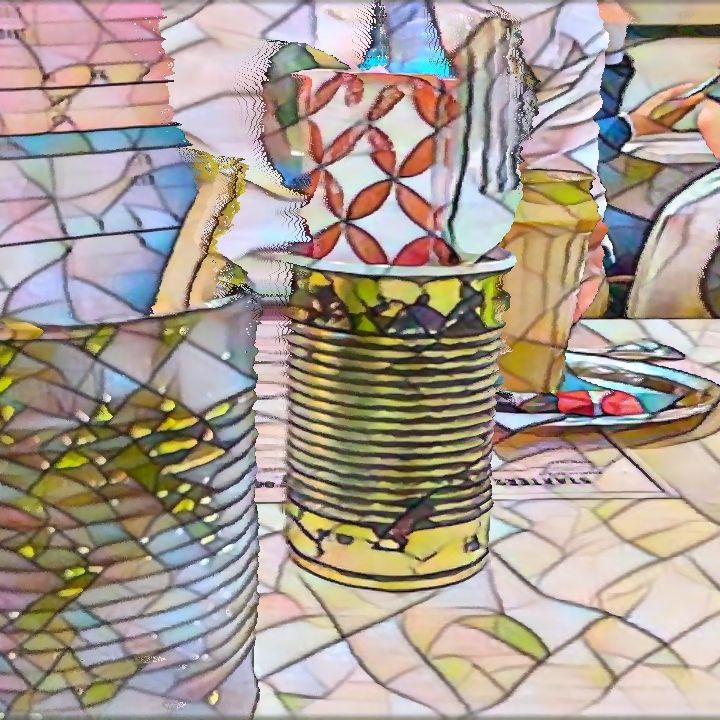} &
    \subfigimg[width = 0.75in]{3}{images/more_results/16v/out_3_result.jpg} &
    \subfigimg[width = 0.75in]{4}{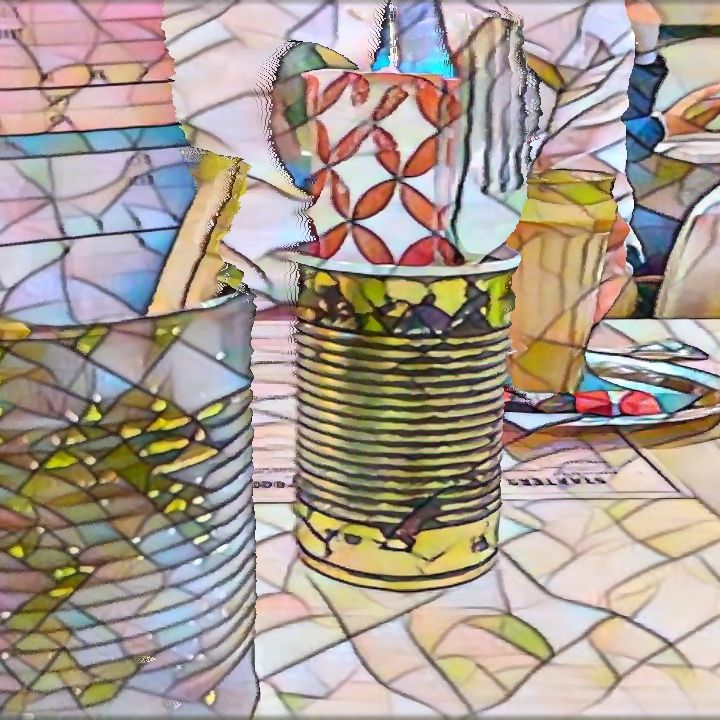}\\
    \subfigimg[width = 0.75in]{5}{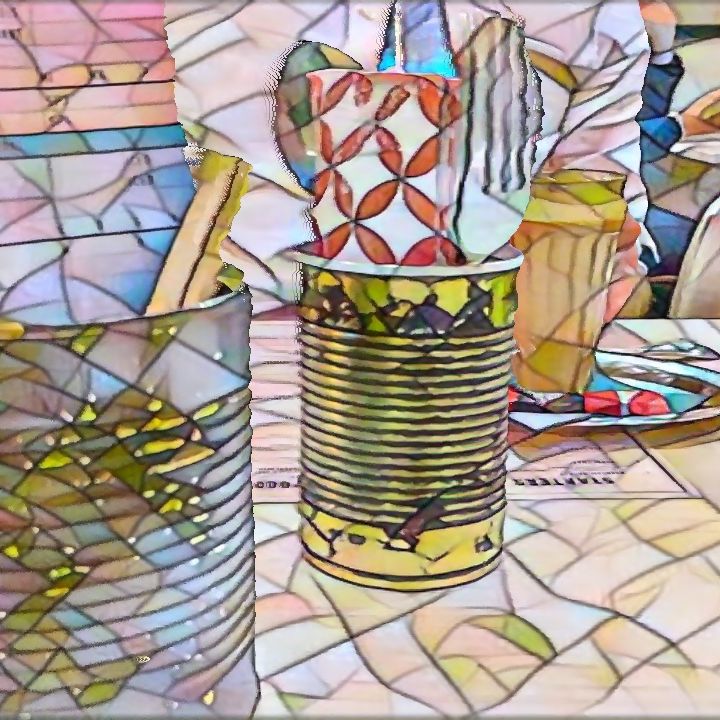} &
    \subfigimg[width = 0.75in]{6}{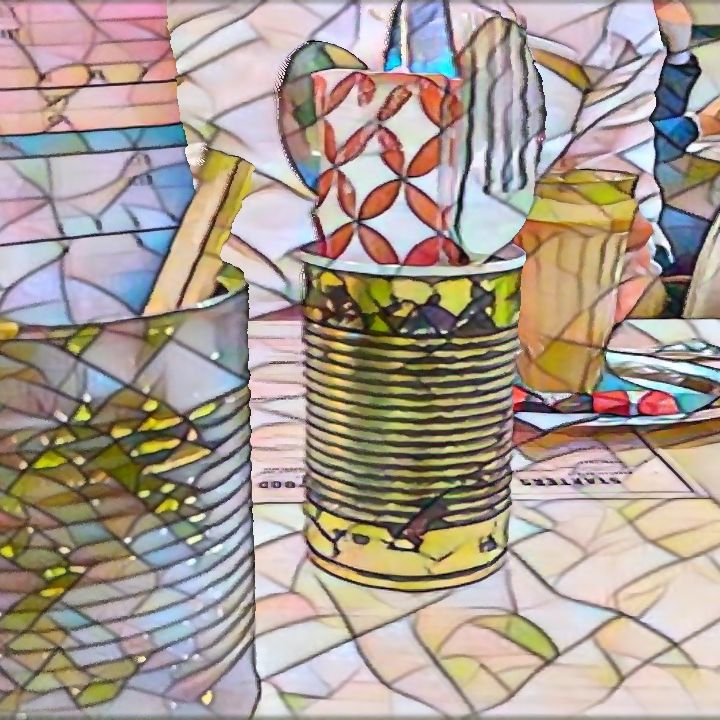} &
    \subfigimg[width = 0.75in]{7}{images/more_results/16v/out_7_result.jpg} &
    \subfigimg[width = 0.75in]{8}{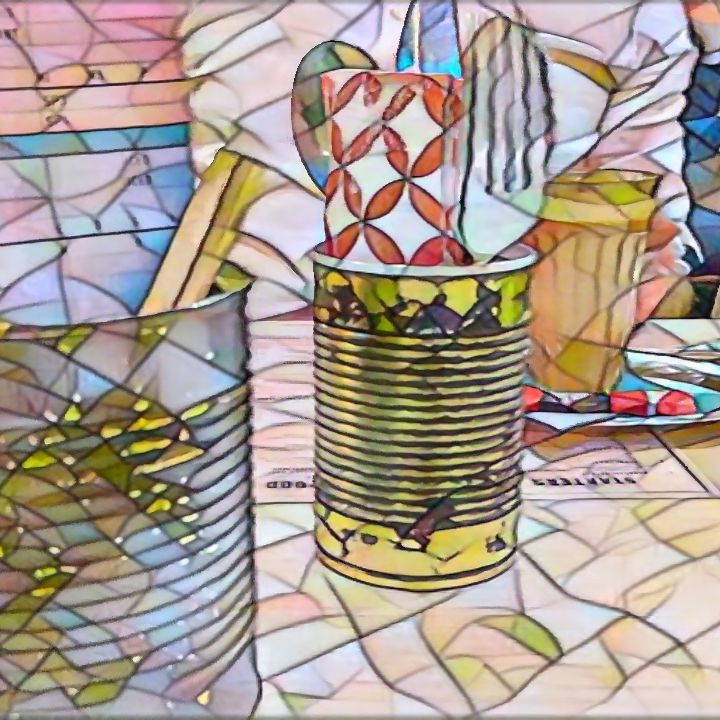}\\
    \subfigimg[width = 0.75in]{9}{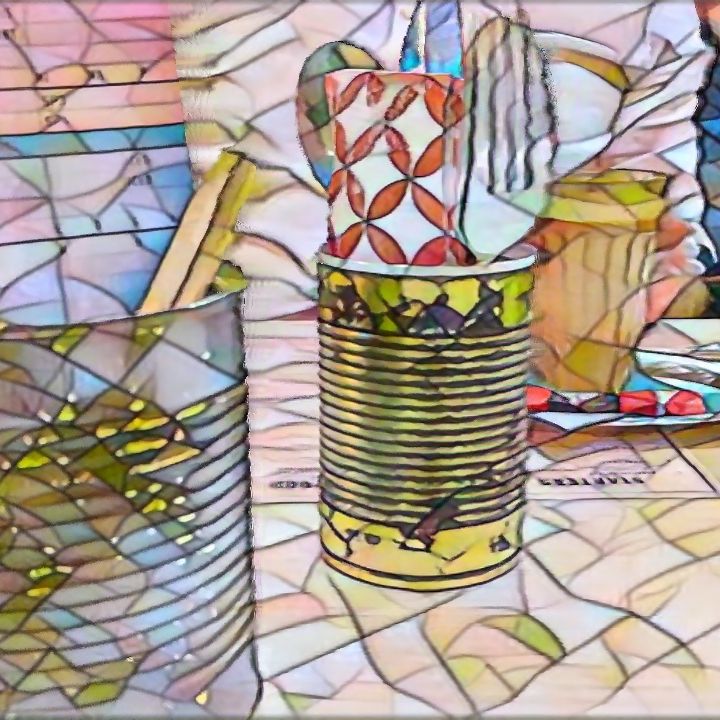} &
    \subfigimgs[width = 0.75in]{10}{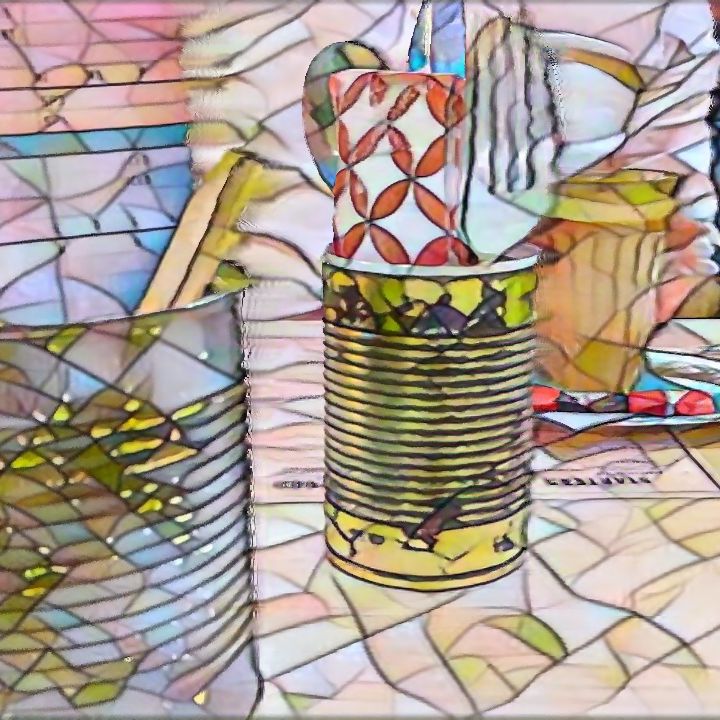} &
    \subfigimgs[width = 0.75in]{11}{images/more_results/16v/out_11_result.jpg} &
    \subfigimgs[width = 0.75in]{12}{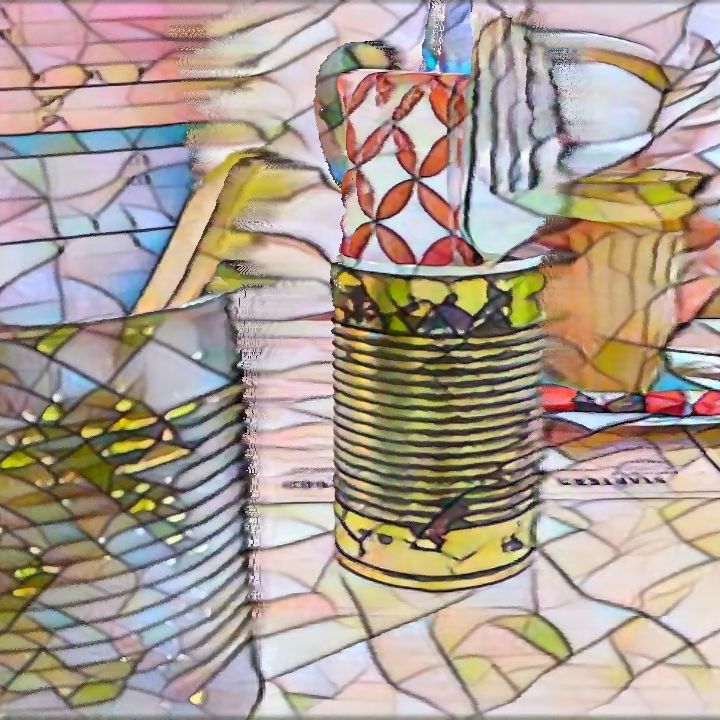}\\
    \subfigimgs[width = 0.75in]{13}{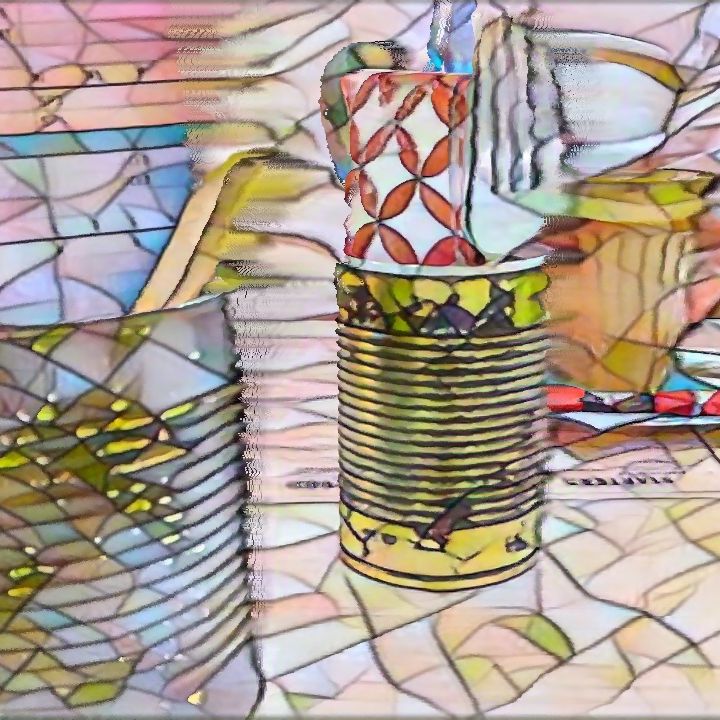} &
    \subfigimgs[width = 0.75in]{14}{images/more_results/16v/out_14_result.jpg} &
    \subfigimgs[width = 0.75in]{15}{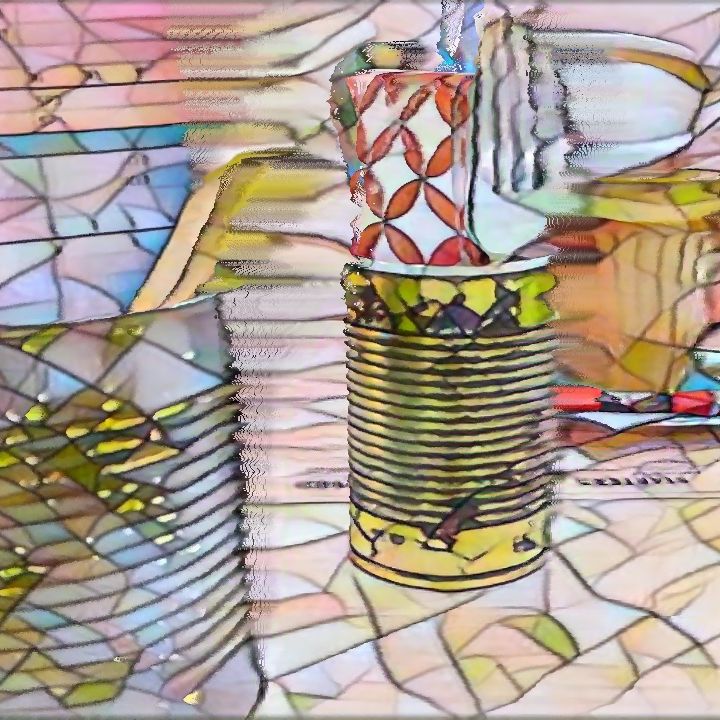} &
    \subfigimgs[width = 0.75in]{16}{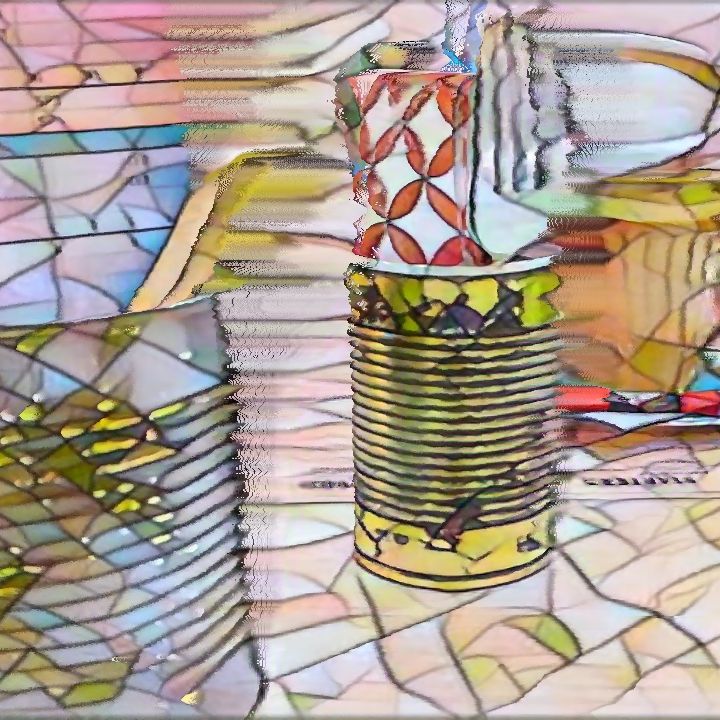}
    \end{tabular}
    \caption{Example results from running our style transfer pipeline and generating 4, 8, and 16 views respectively. The top row of each section shows the original stereo pairs, disparity maps, and style guide used. Each subsequent row shows the generated views from left-most to right-most viewpoint.
    \label{fig:more_results}
}
\end{figure}

\bibliography{references} 
\bibliographystyle{ieeetr}
\end{document}